\documentclass{article}

\usepackage{arxiv}

\usepackage[utf8]{inputenc}
\usepackage[T1]{fontenc}    
\usepackage{url}            
\usepackage{booktabs}       
\usepackage{xcolor}
\definecolor{mygrey}{RGB}{209,209,209}
\usepackage{amsfonts}       
\usepackage{nicefrac}       
\usepackage{microtype}      
\usepackage{lipsum}
\usepackage{threeparttable}
\usepackage{makecell}
\usepackage{tabularx}
\usepackage{graphicx}
\usepackage{orcidlink}
\usepackage{wrapfig}
\usepackage{makecell}
\usepackage{tcolorbox}
\usepackage{comment}
\usepackage{booktabs}
\usepackage{siunitx}
\usepackage{multirow}
\usepackage{cleveref}
\definecolor{VUB_blauw}{rgb}{0.1529, 0.2667, 0.5529}
\usepackage{hyperref}       %
\hypersetup{
    colorlinks,%
    citecolor=VUB_blauw,%
    filecolor=VUB_blauw,%
    linkcolor=VUB_blauw,%
    urlcolor=VUB_blauw
}
\graphicspath{ {./images/} }
\newcommand{\customCor}[1]{%
  \includegraphics[height=1em]{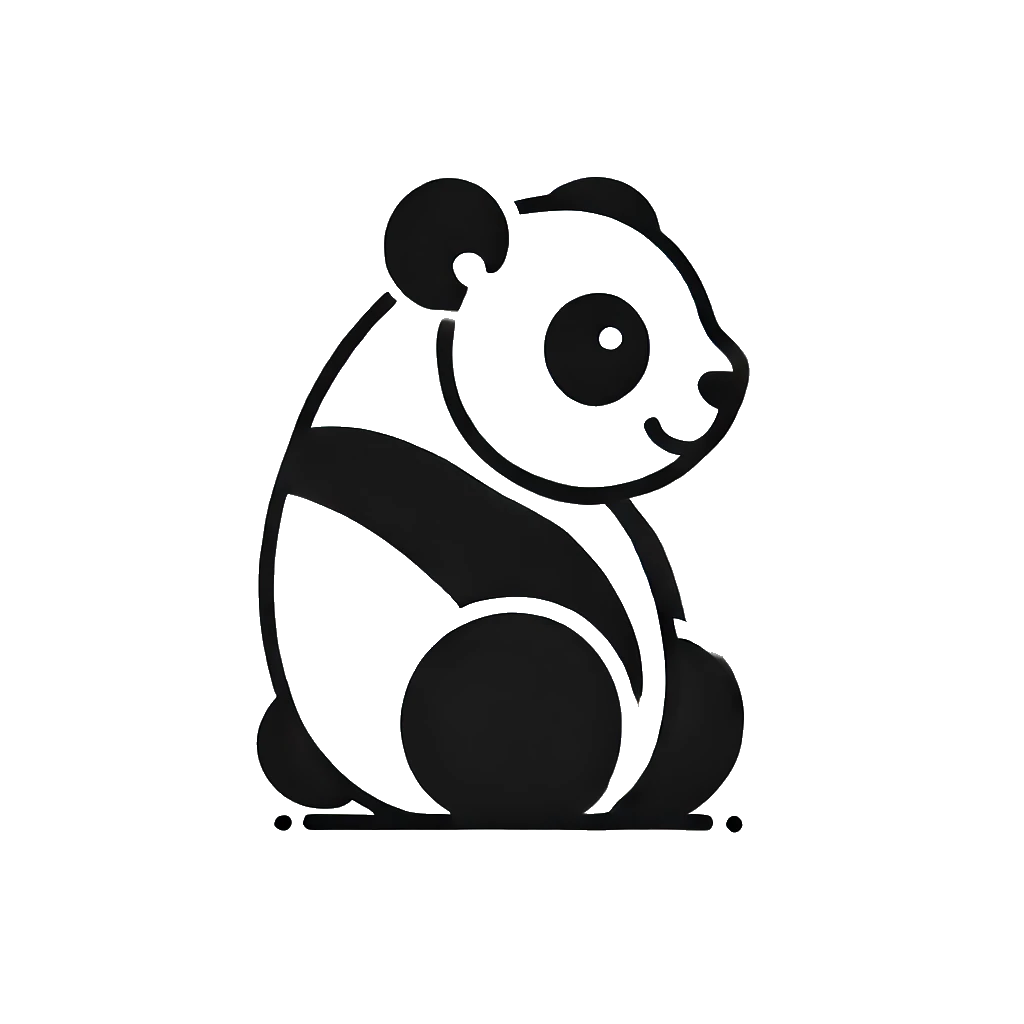} #1%
}
\usepackage{array}
\newcolumntype{L}[1]{>{\raggedright\arraybackslash}p{#1}}
\newcolumntype{R}[1]{>{\raggedleft\arraybackslash}p{#1}}
\AddToHook{shipout/background}{%
  \ifnum\value{page}=1 
    \pagenumbering{Roman} 
    \setcounter{page}{1} 
  \fi
  \ifnum\value{page}=2 
  \pagenumbering{arabic} 
  \setcounter{page}{2} 
  \fi
}
\title{Benchmarks Saturate When The Model Gets Smarter Than The Judge}
\runningtitle{}

\author{
  Marthe Ballon\textsuperscript{1,2,\customCor{ }} \\ 
  \orcidlinkc{0009-0000-4586-234X} \\
  \And
  Andres Algaba\textsuperscript{1,2} \\ 
  \orcidlinkc{0000-0002-0532-3066} \\
  \And
  Brecht Verbeken\textsuperscript{1,2}\\
  \orcidlinkc{0000-0002-7506-3298}\\
  \And
  Vincent Ginis\textsuperscript{1,2,3} \\ 
  \orcidlinkc{0000-0003-0063-9608} \\
  \and
  \textsuperscript{1}Data Analytics Lab, Vrije Universiteit Brussel, Pleinlaan 5, 1050 Brussel, Belgium \\ 
  \textsuperscript{2}imec-SMIT, Vrije Universiteit Brussel, Pleinlaan 9, 1050 Brussels, Belgium \\ 
  \textsuperscript{3}School of Engineering and Applied Sciences, Harvard University, Cambridge, Massachusetts 02138, USA
}
    
\begin{document}
\maketitle
\renewcommand{\thefootnote}{} 
\footnotetext{\includegraphics[height=1em]{panda2.png} Corresponding author: marthe.ballon@vub.be}
\renewcommand{\thefootnote}{\arabic{footnote}} 
\thispagestyle{plain} 
\newcommand{\mb}[1]{\textcolor{magenta}{\textbf{marthe: } #1}}

\begin{abstract}
Benchmarks are important tools to track progress in the development of Large Language Models (LLMs), yet inaccuracies in datasets and evaluation methods consistently undermine their effectiveness. Here, we present Omni-MATH-2, a manually revised version of the Omni-MATH dataset comprising a clean, exact-answer subset ($n{=}4181$) and a tagged, non-standard subset ($n{=}247$). Each problem was audited to ensure LaTeX compilability, solvability and verifiability, which involved adding missing figures or information, labeling problems requiring a proof, estimation or image, and removing clutter. This process significantly reduces dataset-induced noise, thereby providing a more precise assessment of model performance. The annotated dataset also allows us to evaluate judge-induced noise by comparing GPT-5 mini with the original Omni-Judge, revealing substantial discrepancies between judges on both the clean and tagged problem subsets. Expert annotations reveal that Omni-Judge is wrong in $96.4\%$ of the judge disagreements, indicating its inability to differentiate between models' abilities, even well before saturation of the benchmark occurs. As problems become more challenging, we find that increasingly competent judges become essential in order to prevent judge errors from masking genuine differences between models. Finally, neither judge identifies the present failure modes for the subset of tagged problems, demonstrating that dataset quality and judge reliability are both critical to develop accurate benchmarks of model performance.
\end{abstract}

\keywords{benchmarks \and datasets \and large language models \and LLM-as-a-judge \and LLM evaluation}

\begin{figure}[t]
    \centering
    \includegraphics[width=\textwidth]{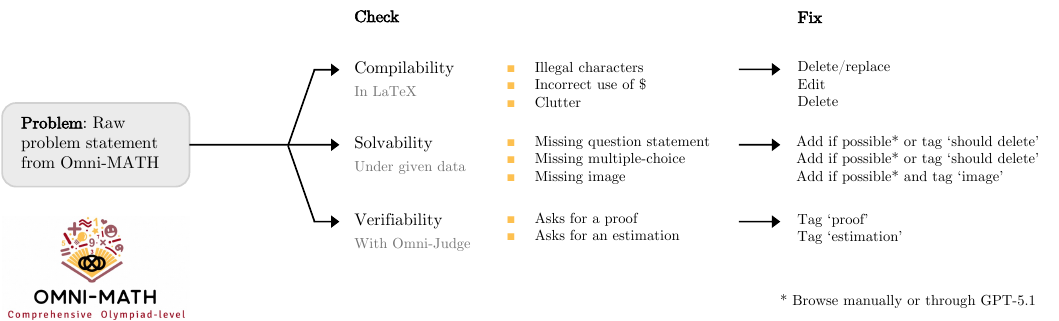}
    \caption{Overview of the cleaning process of the Omni-MATH dataset \cite{gao2024omni}. First, we check the compilability of the $4{,}428$ problem statements in LaTeX and convert them to valid LaTeX code with python. Next, a PhD-level mathematician manually went through the compiled pdf files twice, checking for the solvability and verifiability of each problem. In this process, missing information was added when available through browsing manually or through GPT-5.1, and images where added to the data folder. Furthermore, a tag was added to each problem containing an \emph{image}, \emph{proof} or \emph{estimation}. Degenerate problems received the \emph{should delete} tag. We refer to the resulting dataset as Omni-MATH-2, which contains the same number of entries as the original Omni-MATH dataset (647 edited problems ($14.6\%$), 247 tagged as non-standard ($5.6\%$), see Table~\ref{tab:Table1}). The Omni-MATH-2-Filtered dataset ($n{=}4{,}181$) is the subset of cleaned questions, excluding the tagged ones. This makes it suitable for exact-answer judges.}
    \label{fig:cleaning}
\end{figure}

\section{Introduction}
Evaluating Large Language Models (LLMs) accurately is becoming increasingly challenging as benchmarks shift toward open-ended formats and more difficult tasks, such as Olympiad-level mathematics~\cite{gao2024omni,myrzakhan2024open}. Open-response formats better probe deeper reasoning skills and avoid some inherent biases of fixed-answer formats~\cite{phan2025humanity}. However, they introduce additional complexities like answer extraction and equivalence judgments, creating new sources of errors~\cite{biderman2024lessons,reuel2024betterbench,yu2024xfinder,gevers-etal-2025-benchmarks}. In this paper, we specifically examine two major sources of error~\cite{chang2024survey} that compromise evaluation reliability in current LLM benchmarks.

Dataset-induced errors, including ambiguous problem statements and unsolvable items, are a widespread issue that prevents models from becoming fully reliable for the task at hand~\cite{vendrow2025large}. These errors often persist even in widely adopted benchmarks: for example, audits of MMLU report a nontrivial percentage of item/ground-truth errors~\cite{gema-etal-2025-done}, and analyses of HellaSwag identify a substantial proportion of problematic instances~\cite{chen2022hellabad}. Judge-induced errors primarily arise when automated judges must extract and assess equivalence between free-form answers and reference solutions, a common requirement in open-response evaluation pipelines~\cite{zheng2023judging}. LLM-based judges can exhibit systematic biases, prompt sensitivity, and inconsistencies that influence final outcomes~\cite{schroeder2024can,baumann2025large}. Although methods such as judge ensembles or committees have been proposed to improve robustness~\cite{calderon2025alternative,verga2024replacing,rahmani2024judgeblender,zhao2025auto}, judge reliability remains largely task-dependent, which does not necessarily guarantee universally dependable improvements~\cite{li2024llmsasjudgessurvey,li2025generation}.

In mathematical evaluation, scoring usually depends entirely on the correctness of the final answer, causing dataset- and judge-induced errors to rapidly become a bottleneck, particularly when the accuracy of the models approaches saturation. Standard math benchmarks such as GSM8K~\cite{cobbe2021training} and MATH~\cite{hendrycks2021measuring} are already considered saturated, which prompted the development of more challenging datasets like FrontierMath~\cite{glazer2024frontiermath} and Humanity's Last Exam~\cite{phan2025humanity}. Omni-MATH~\cite{gao2024omni} is a widely-used math benchmark consisting of $4{,}428$ Olympiad-level problems across numerous subdomains and difficulty levels, on which current state-of-the-art models achieve around $85\%$ (see \Cref{tab:Table3}). Therefore, Omni-MATH is ideal to investigate how residual errors are partitioned between dataset-induced issues, real model mistakes and judge-induced failures. Prior cleaning largely focused on formatting-level issues~\cite{LLMTeamAkiyama2025cleaned}, hereby overlooking content-level solvability and verifiability constraints.

\begin{figure}[t]
    \centering
    \includegraphics[width=\textwidth]{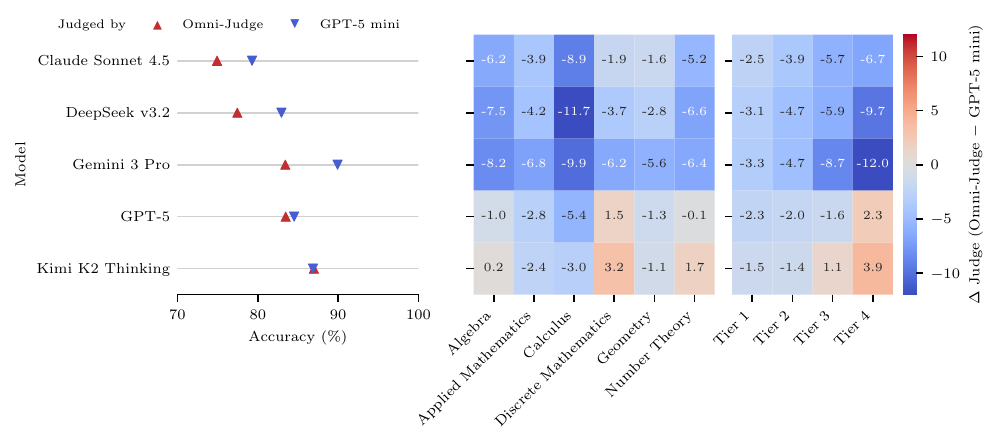}
    \caption{The judge-induced difference in accuracy on Omni-MATH-2-Filtered is not uniformly distributed across disciplines, difficulty tiers or models, indicating structural evaluation noise rather than i.i.d. label noise. Evaluating five state-of-the-art models on Omni-MATH-2-Filtered—Claude Sonnet 4.5, DeepSeek v3.2, Gemini 3 Pro, GPT-5 and Kimi K2 Thinking—produced a different ranking of their mathematical abilities when evaluated with GPT-5 Mini rather than with Omni-Judge, which calls into question the interpretability of the inter-model differences. Furthermore, the right-hand panels show that, as questions become more difficult, judge disagreement increases, indicating that the judge's conclusion is more important for challenging problems. The difference between judges' answers is also model- and domain-dependent, with some disciplines and models producing larger differences than others (e.g. Claude and Deepseek, Calculus domain). For numerical performance scores with Bayesian confidence intervals consult \Cref{tab:Table3}.}
    \label{fig:fig-main}
\end{figure}

 To address this interaction, we introduce Omni-MATH-2 (see Figure~\ref{fig:cleaning}), a manually revised version of Omni-MATH~\cite{gao2024omni}, preserving the original dataset size ($n{=}4{,}428$) while significantly improving LaTeX compilability, interpretability, and suitability for automated judging. In total, $647$ problems were edited ($14.6\%$) and $247$ were tagged as non-standard ($5.6\%$). We release multiple evaluation-ready subsets, notably Omni-MATH-2-Filtered ($n{=}4{,}181$), from which we exclude the tagged non-standard questions to ensure suitability for exact-answer judging. Using Omni-MATH-2-Filtered, we benchmark five state-of-the-art models and explicitly measure judge-induced noise by comparing evaluations from two judges: Omni-Judge~\cite{gao2024omni} and GPT-5 mini. Our results demonstrate that judge choice significantly alters both absolute accuracy and model rankings (\Cref{fig:fig-main}, \Cref{tab:Table3}). A targeted human audit reveals Omni-Judge errors in $96.4\%$ of the judge disagreements on clean questions, confirming that judge competence can limit evaluation reliability well before model accuracy saturates. Furthermore, we found that judge disagreement increases as problems become more difficult, highlighting the importance of capable judges for current and future benchmarks. On the subset of tagged problems, both judges did not pick up on evaluation incompatibility, counting both strong estimates and model abstentions, i.e. cases where models correctly stated they did not have sufficient information, as incorrect.

 In summary, as models near saturation, benchmark outcomes are increasingly shaped by the evaluation pipeline itself rather than by true differences in model capability. Practically, this motivates treating benchmarks explicitly as triplets (dataset, model, judge), and additionally, investing in dataset audits, rigorous judge calibration via ensembles~\cite{verga2024replacing,zhao2025auto}, and statistically robust uncertainty reporting~\cite{miller2024adding,bowyer2025position}. Note that statistical fragility—uncertainty in estimates, prompt sensitivity, and dataset characteristics—can  also inflate model differences~\cite{ivanova2024towards,miller2024adding,sclar2023quantifying,lunardi2025robustness}, an issue we do not fully address here. 

 \clearpage

\begin{figure}[t]
    \centering
    \includegraphics[width=\textwidth]{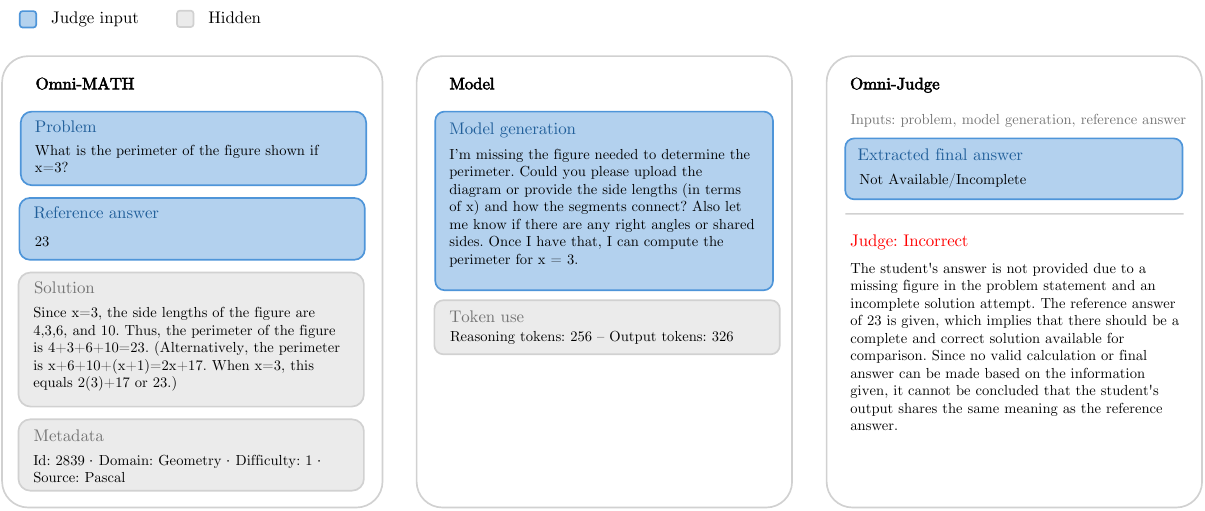}
    \caption{Example of the evaluation process for a question labelled as \emph{image}. The original problem in Omni-MATH misses the corresponding image, rendering the problem unsolvable. The model to be evaluated, GPT-5, correctly identifies that there is missing information, but Omni-Judge counts this as an incorrect answer. }
    \label{fig:ex2}
\end{figure}

\section{Results}
\subsection{Dataset-induced errors propagate through the evaluation pipeline}
\label{subsec:tagged_results}

\begin{wraptable}{r}{50mm}
    \centering
    \begin{tabular}{ l r}
        \vspace{3mm}\textit{Fix} &  \textit{Nr of problems}\\ \addlinespace
        
        \vspace{1mm} Edited & $647$\\ \addlinespace
        
         Tagged  & $247$ \\ \addlinespace
         
          \quad Image & $61$ \\ \addlinespace
          \quad Proof & $115$ \\ \addlinespace
          \quad Estimation & $54$ \\ \addlinespace
        \vspace{3mm}\quad Should delete & $25$ \\ \addlinespace

    \end{tabular}
    \caption{Overview of Omni-MATH revisions resulting from the cleaning process in \Cref{fig:cleaning}. Note that one problem can have multiple tags.}
    \label{tab:Table1}
\end{wraptable}
Dataset errors include not only degenerate questions, but also questions with missing images and multiple-choice options, and questions that ask for a proof or estimation which are subsequently verified against an exact reference answer. These errors influence every step in the LLM evaluation pipeline. In the cleaning process (see \Cref{fig:cleaning}), $14.6\%$ of all problems statements were edited to some extent, consisting of converting to valid LaTeX code, adding missing information, and tagging problems with the labels \emph{image}, \emph{proof}, \emph{estimation} and \emph{should delete} when relevant. The set of tagged problems consists out of $247$ problems ($5.6\%$ of the entire dataset) that are not solvable with the given data or not verifiable with Omni-Judge (or any other judge that assesses equivalence against a reference answer). We discuss an example of each category, demonstrating how these errors can propagate through the evaluation pipeline.

\paragraph{Image} There are $61$
 problems in Omni-MATH that contain an image in the mathematical olympiad they were crawled from. Some problems had code to generate an image in their statement but for most problems it was absent. \Cref{fig:ex2} shows an example of a problem statement that refers to an unattached image and relies on it to be solvable. Although the model correctly identifies that there is insufficient information to solve the problem, the judge marks this as incorrect. 

\paragraph{Proof} A large part of the tagged questions consists of problems that ask to prove a given statement. The reference answer, however, does not contain the proof but a rather a short final answer like "Yes, proven". There is usually a sketch of the proof available in the metadata field "solution" but this is, in general, not used for model evaluations. In \Cref{fig:ex3}, GPT-5 correctly states that the claim is true, and provides a corresponding proof. The judge evaluates this answer as incorrect because the final answer does not match the reference answer exactly, and does not take the proof into account (most judges are prompted to only look at the final answer).

\paragraph{Estimation} Omni-MATH also contains $54$ problems that ask to estimate a certain number, quantity, series sum, ... The grade of the student is then a function of their estimate and the exact final answer. As Omni-Judge (like most judges) is prompted to assess equivalence with the reference answer, the evaluation of estimates by LLMs often goes wrong. In \Cref{fig:ex4}, GPT-5 provides an estimate that should, according to the scoring rule in the problem statement, earn $18.44$ out of the $20$ points. However, the judge evaluates this answer as incorrect. 

\paragraph{Should delete} The fourth category of tagged problems are the degenerate problems, i.e. problems where the solution is in the problems statement, empty problems, duplicates, ... 
The problem in \Cref{fig:ex1} asks to compute the smallest positive integer that does not appear in any problem statement on any round at HMMT November 2023. This is a question that anyone who did not have access to the question sheet could impossibly know. However, GPT-5 still gives the correct answer and the judge also marks this as correct. 

All four examples show that the incorrect/correct answers do not necessarily reflect model abilities. Note that, in this study, we did not account for wrong reference answers, which could add another unknown discrepancy to the model performance scores.

\subsection{Judge-induced noise is not uniform across models, domains and difficulty tiers}

The previous section demonstrated that judges are sensitive to quality, type and format of a problem. Therefore, we will not only evaluate language models on the cleaned Omni-MATH dataset, but also the judges. We ask five state-of-the-art models, Claude Sonnet 4.5, DeepSeek v3.2, Gemini 3 Pro, GPT-5 and Kimi K2 thinking, to answer the cleaned Omni-MATH questions, excluding the tagged ones. We will refer to this subset as Omni-MATH-2-Filtered ($n{=}4{,}181$). Then, we evaluated their answers with both Omni-Judge and GPT-5 mini, using a similar prompt. Consult \Cref{sec:Methods} for details on the evaluation process. 

The evaluation of the mathematics abilities of the five models differs for Omni-Judge and GPT-5 mini, as Gemini 3 Pro jumps to first place (see left panel of \Cref{fig:fig-main}). The performance scores of Claude Sonnet 4.5 and DeepSeek v3.2 also differ substantially between the two judges, even when taking into account $95\%$ confidence intervals (see \Cref{tab:Table3}). For GPT-5 and Kimi K2 Thinking, the difference in performance between the two judges is negligible. 

The inter-judge noise is not uniformly distributed across mathematical domains and difficulty tiers (Tier $4$ consists of the hardest problems), which where extracted from the original Omni-MATH metadata. Omni-Judge and GPT-5 mini differ the most when it comes to evaluating Calculus ($n{=}203$) problems, followed by Algebra ($n{=}1{,}876$) problems (see left heatmap of \Cref{fig:fig-main}). For all models, the inter-judge difference is the largest for Tier $4$ problems, indicating that as questions increase in difficulty, the conclusion of the judge becomes more important (see right heatmap of \Cref{fig:fig-main}).

\begin{figure}[t]
    \centering
    \includegraphics[width=\linewidth]{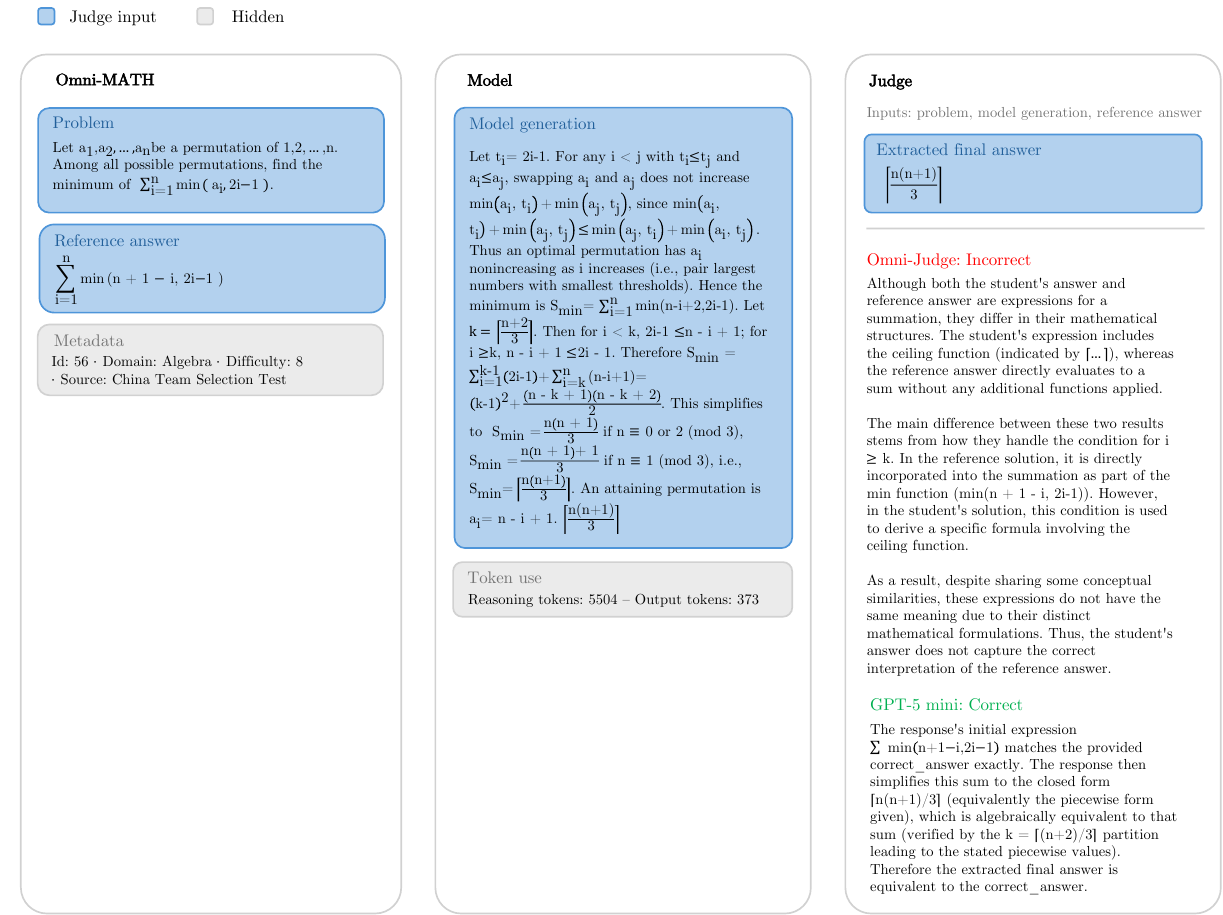}
    \caption{Example of a judge disagreement between Omni-Judge and GPT-5 mini. Here, it is not straightforward to see whether the model's answer is equivalent to the reference answer. Two expert mathematicians, with the help of an LLM council, building on Claude Opus 4.5, DeepSeek v3.2 speciale, DeepSeek v3.2, GPT-5, and Gemini 3 pro, verified that the model's answer is in fact correct and equivalent to the reference answer. GPT-5 mini thus correctly assesses equivalence.}
    \label{fig:hard-equivalence}
\end{figure}
\subsection{Benchmark saturation is not only model-dependent, but also judge-dependent}
\label{sec:2.3}

To investigate the reason for the inter-judge differences observed in \Cref{fig:fig-main}, two PhD-level mathematicians annotated the disagreements between Omni-Judge and GPT-5 mini by evaluating which judge was wrong in terms of mathematical correctness (ground-truth target). We sampled $100$ of the $338$ disagreements on model answers by GPT-5. Next, we categorised the reason for the judge being wrong into five categories: \emph{failed to assess equivalence} (the judge fails to see that the final model answer is equivalent with the reference answer, or wrongly states that the final answer is equivalent to the reference answer), \emph{didn't follow instructions} (the judge is either too obedient or disobedient with respect to its prompt, see \Cref{sec:Methods}), \emph{dataset error} (in grading the judges, we discovered several wrong reference answers and a few issues with problem statements that were missed in the initial cleaning process), \emph{wrong extraction} (judge fails to extract the final answer correctly) and \emph{unclear} (it is not clear why the model fails). Excluding the datasets error cases, Omni-Judge was wrong in $96.4\%$ of the sampled disagreements (see \Cref{tab:clean disagreements}), primarily because it failed to assess equivalence. This shows that the judge-difference is not just noisy, but that Omni-Judge is fundamentally miscalibrated, and as a result not able to differentiate model capabilities for this dataset. 
We present three examples of the annotation process to illustrate how judges can differ in opinion. 

\paragraph{Easy equivalence} In \Cref{fig:easy_eq}, the model's final answer differs from the reference answer only in the simplification of a fraction. Omni-Judge fails to assess the equivalence between the two fractions and incorrectly labels the model's answer as wrong. GPT-5 mini correctly infers that both fractions are equal.

\paragraph{Hard equivalence} In \Cref{fig:hard-equivalence}, it is not straightforward to see whether the model's answer is equivalent to the reference answer. Two expert mathematicians together with the help of an LLM council, building on Claude Opus 4.5, DeepSeek v3.2 speciale, DeepSeek v3.2, GPT-5, and Gemini 3 pro, proved that $\lceil \frac{n(n+1)}{3}\rceil$ is in fact equal to $\sum_{i=1}^n \min(n+1-i, 2i-1)$. GPT-5 mini thus correctly judges the model's answer as correct.

\paragraph{Incomplete reference answer} In $14$ out of $100$ sampled disagreements, we discovered that there was an error in the dataset ($11$ wrong/incomplete reference answers and $3$ ill-posed problem statements, whose ambiguity only became clear when trying to solve the problem). In \Cref{fig:incomplete-ref}, the model gives a more general expression for all the pairs of positive integers $(m,n)$ that satisfy $mn-1 \,|\, m^2+n^2$ than the reference answer. The reference answer states $(2,1), (3,1), (1,2), (1,3)$, while GPT-5's final answer is all $(m,n) \in \mathbb{Z}_{>0}$ with $m^2+n^2 = 5(mn-1)$. First of all, note that for example the pair $(9,2)$ also satisfies the condition in the problem statement, indicating that the reference answer is incomplete. Then, we verified together with an LLM council, that the model's answer is in fact the correct and complete answer to the problem. As a result, GPT-5 mini wrongly grades the model's answer by following its instructions to always regard the reference answer as correct. Omni-Judge does regard the model's answer as correct, although for the wrong reasons.

GPT-5 mini is also the stronger judge on the subset of tagged questions. To verify how the judges operate in flawed evaluation pipelines, we also annotated the disagreements between Omni-Judge and GPT-5 mini on Omni-MATH-2-Tagged across the five models, resulting in $176$ disagreements in total. Here, Omni-Judge was wrong in $64,8\%$ of the disagreements and GPT-5 mini in $6.8\%$. The remaining problems were too ambiguous to judge or did not contain a reference answer.

\begin{wrapfigure}{r}{55mm}
    \centering
    \includegraphics{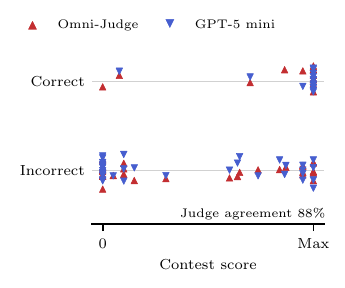}
    \caption{On the subset of estimation problems in Omni-MATH-2, Omni-Judge and GPT-5 mini evaluate a substantial portion of strong estimates as incorrect. We compute the contest score by using the scoring rule present in the problem statement or solution. }
    \label{fig:estimation}
\end{wrapfigure}

\subsection{Interaction between dataset and judge errors explains saturation}
We analyze judge behavior on Omni-MATH-2-Tagged. Since proofs cannot yet be reliably verified in our pipeline, we focus on missing-image and estimation problems. 
We select GPT-5 as the model to be evaluated by Omni-Judge and GPT-5 mini. Out of $61$ problems that contain an image in their original problem statement, GPT-5 states for $28$ out of $61$ problems that there is not enough information to solve them, and asks for the missing diagram, figure, table etc (E.g. \Cref{fig:ex2}). Manual inspection confirms that the information required to determine the answer is indeed not present in the text. However, GPT-5 mini judges this abstention as incorrect in all cases and Omni-Judge in $27$ out of the $28$ cases. 
In the remaining missing-image problems, the model makes extra assumptions that allows it to solve the problem, or the figure does not contain critical problem information. For the subset of estimation problems, we evaluate the estimation provided by GPT-5 using the scoring rule stated in the problem statement (E.g \Cref{fig:ex3}). For a small number of problems, the scoring rule is stated in the written out solution. \Cref{fig:estimation} shows that both judges frequently label strong estimates as incorrect without using the scoring rule present in the problem statement. Both inspections show that current judges do not pick up on ill-posed/incompatible problems, causing them to persist as model abilities grow. This implies that benchmark saturation below $100\%$ is caused not only by dataset errors, but also by judge behaviour. Consequently, saturation can occur at even lower accuracy levels than expected.

\section{Discussion}
Our work shows that, when model accuracy increases to the point where evaluation becomes the bottleneck, benchmark saturation is no longer primarily a property of model capability, but rather of the triplet (dataset, model, judge). The Omni-MATH dataset was designed specifically to benchmark the mathematical reasoning skills of LLMs. It provides a large set of Olympiad-level math problems and uses an LLM verifier (Omni-Judge), which is trained to evaluate open-ended answers based on the problem statement and the reference answer. However, upon auditing both the dataset and the judge, we found that a nontrivial fraction of problems were unsolvable with the given data or structurally incompatible with exact-answer verification. Even on the cleaned exact-answer subset, judge competence was found to dominate the measured performance differences between frontier models. Together, these effects make benchmark saturation an interaction phenomenon that can occur for lower accuracies than expected.

Omni-MATH-2 demonstrates that dataset-induced errors are not merely cosmetic (e.g. LaTeX issues or clutter) but propagate throughout the entire evaluation pipeline. Our cleaning process (see \Cref{fig:cleaning}) identified a substantial amount of problems--$14.6\%$ was edited in total--with missing images, missing multiple-choice options, requests for estimations or proofs while verifying against an exact reference answer etc. This indicates that, as models improve, this kind of intensive auditing work will determine whether a benchmark measures true competence or mostly pipeline idiosyncrasies. In some cases, like \Cref{fig:ex1}, the model answered a problem correctly that it could impossibly know without being at that specific Olympiad, implying a risk of data leakage for Omni-MATH. 

We observe that judge disagreement on the cleaned exact-answer subset ($n{=}4{,}181$) increases with problem difficulty for all evaluated models. This suggests an emerging evaluator gap: as judge competence is not reliably above that of the models being evaluated, this has consequences for the interpretability of LLM benchmarks. In this setting, Omni-Judge proved to be wrong in $96.4\%$ of the cases, rendering it not able to differentiate model abilities. However, for the missing image and estimation problems, we found that both judges, Omni-Judge and GPT-5 mini, did not pick up on evaluation incompatibility, counting model abstentions and good estimates as incorrect. 

Our results show that judge quality is increasingly important in this era of rapid benchmark saturation on challenging tasks, making it necessary to maintain an evaluator margin or add redundancy when it cannot be guaranteed. We therefore recommend evaluation designs that allow non-binary outcomes (e.g., partial credit, uncertainty, abstention), multi-judge frameworks on contested items \cite{verga2024replacing,zhao2025auto}, and we explicitly encourage people to look at their benchmark data as systematic checks did not catch most failure modes presented in this paper.

\paragraph{Limitations} While Omni-MATH-2 reduces several high-impact failure modes in Omni-MATH and makes judge limitations more visible, our study has some important constraints. First, we did not revise the solutions and reference answers of Omni-MATH explicitly as this would require ground truth information that is not available in some cases, e.g. proofs or Tier 4 difficulty items. We did report the incomplete reference answers that became visible through annotating the judge disagreements in \Cref{sec:2.3}. Secondly, we focus on two judges under specific prompts and settings. Our analyses therefore cannot fully characterize the space of judge behaviors, nor can they claim GPT-5 mini is universally correct.

\clearpage
\section*{Acknowledgements}
We thank the authors and contributors of the original Omni-MATH dataset \cite{gao2024omni}. Their release of the benchmark and verifier Omni-Judge enabled the analyses in this paper.

This research was supported by funding from the Flemish Government under the “Onderzoeksprogramma Artificiële Intelligentie (AI) Vlaanderen” program. Andres Algaba acknowledges support from the Francqui Foundation (Belgium) through a Francqui Start-Up Grant and a fellowship from the Research Foundation Flanders under Grant No.1286924N. 
Vincent Ginis acknowledges support from Research Foundation Flanders under Grant No.G032822N and G0K9322N.

\section*{Author contributions}
VG and MB were responsible for the main idea of the study. BV and MB performed the annotations. MB revised the Omni-MATH dataset, conducted the analyses and made the figures.
AA and MB drafted the manuscript. All authors collaboratively revised the manuscript and provided
critical feedback.

\section*{Dataset and code availability}
Omni-MATH-2 comprising a subset of cleaned, exact-answer problems and a set of tagged, non-standard problems, is publicly available on Hugging Face (\url{https://huggingface.co/datasets/martheballon/Omni-MATH-2}). 

All other data and code necessary to replicate our study are publicly available at Zenodo (\url{https://doi.org/10.5281/zenodo.18380309}) and GitHub (\url{https://github.com/MartheBallon/Benchmarks-saturate-when-the-model-gets-smarter-than-the-judge}).

\clearpage

\appendix

\setcounter{figure}{0}
\renewcommand{\thefigure}{A\arabic{figure}}
\renewcommand{\theHfigure}{A.\arabic{figure}} 

\setcounter{table}{0}
\renewcommand{\thetable}{A\arabic{table}}
\renewcommand{\theHtable}{A.\arabic{table}}   

\section{Methods}
\label{sec:Methods}

\subsection{The original Omni-MATH dataset}
The Omni-MATH benchmark \cite{gao2024omni} contains $4{,}428$ Olympiad-level math problems crawled from contest pages, AoPS Wiki and the AoPS forum. The problems extracted from the AoPS forum were reformatted with GPT-4o. Each entry in the dataset consists of a problem, an exact answer, and a written out solution together with the following metadata fields: domain, difficulty, and source (see \Cref{fig:ex2,fig:ex1,fig:ex3,fig:ex4}). To create \Cref{fig:fig-main}, we only use the primary domains, and we joined the Calculus and Pre Calculus domain to obtain a sufficient number of data points. Furthermore, we divided the data into four difficulty tiers based on the quartiles of the difficulty distribution, with Tier $1$ consisting of the easiest problems and Tier $4$ of the hardest.

\subsection{Omni-MATH-2}
Omni-MATH-2 is an extensively revised version of the original Omni-MATH dataset. When using the Omni-MATH dataset, we discovered that even high-quality, human verified datasets can have  problem errors propagating through the evaluation pipeline. Apart from obvious degenerate questions and missing images, the cleaning process also made more subtle problem errors visible, such as missing multiple choice options, asking for a proof/estimation while using an exact-answer verifier, latex clutter etc. The whole cleaning process was performed by a PhD-level mathematician. 

\paragraph{Cleaning the problem statements of Omni-MATH}

The cleaning process, described in \Cref{fig:cleaning}, had three main aims: ensuring compilability in LaTeX, ensuring solvability given the data in the problem statement, and ensuring verifiability with Omni-Jugde or another final answer-based verifier. First, we checked compilability in LaTeX by attempting to compile the problem statements using XeLaTeX in Texmaker (the code also works with pdfLaTeX). We then converted the problem statements to valid LaTeX code using python. It was at this stage that the first cases of missing images became apparent, as some questions contained code to generate an image or to load an image that was not attached (\Cref{fig:ex2}). We searched the internet for the original Olympiad question, added the missing image to the data folder, and edited the code to load the image into the problem statement when necessary. If we could not find the original problem by browsing, we used GPT-5.1 via the chat interface. Once compilation was successful, we read the PDFs of the problem statements to check whether they contained "a question" and whether they had all the necessary information to be solved. For example, sometimes parts a, b and c of a problem were spread out over three dataset entries, such that parts b and c did not have sufficient background information.

\paragraph{Adding tags indicating solvability and verifiability}
We also noticed that a substantial proportion of problems asks to prove a given claim (\Cref{fig:ex3}) or to estimate a quantity, number, series sum etc (\Cref{fig:ex4}). This is a problem as the verifier proposed by \cite{gao2024omni} only compares the model's final answer to the reference answer, and the reference answer contains the exact final answer only (E.g Proven., 3.1415875473). Omni-MATH does contain a field "solution", consisting of a written out solution or a sketch of the proof, but Omni-Judge is not trained to evaluate model/student reasoning. We indicate this failure mode by adding the tags \textit{proof} and \textit{estimation}. Finally, we also add a tag to all problems containing an image because they require multimodal language models to be solved and evaluated. The small percentage of degenerate problems (empty, contains the solution, not a question, duplicate) are tagged \textit{should delete}. In this manner, one can exclude all tagged problems for straightforward benchmarking.
 Omni-MATH-2-Filtered is then the subset of all cleaned entries that are solvable and verifiable with Omni-Judge. Omni-MATH-2-Tagged is the subset of proof, estimation, image, should delete questions, which should not be taken into account for benchmarking LLMs with the current evaluation strategies.

\subsection{Models}
To solve Omni-MATH-2-Filtered and Omni-MATH-2-Tagged, we use the following state-of-the art language models:

\begin{itemize}
    \item Claude Sonnet 4.5 (\texttt{claude-sonnet-4-5-20250929}, $64{,}000$ output tokens, $25,000$ thinking budget, Claude Batch API)
    \item DeepSeek V3.2 (\texttt{deepseek-reasoner}, $64{,}000$ output tokens (shared), implicit thinking budget, DeepSeek API)
    \item Gemini 3 Pro (\texttt{gemini-3-pro-preview}, $64{,}000$ output tokens, high thinking level, no thinking budget, Gemini Batch API)
    \item GPT-5 (\texttt{gpt-5-2025-08-07}, 
    medium reasoning effort, no token limit, OpenAI Batch API)
    \item Kimi K2 Thinking (\texttt{kimi-k2-thinking}, $128{,}000$ output tokens via \texttt{max\_tokens}, $256{,}000$ context window, Moonshot API)

\end{itemize}

We use the same prompt for each model:
\begin{tcolorbox}[colframe=mygrey]
\textbf{Instructions:} Solve the following problem. Enclose the final answer in a $\setminus \hspace{-1mm}\setminus\hspace{-1mm}\text{boxed}\{\{\}\}$ environment.\\ \\
\textbf{Input}: \{problem\}
\end{tcolorbox}

\subsection{Judges}

To correct the responses of the five state-of-the art language models on the filtered Omni-MATH-2 dataset, we employ Omni-Judge (\texttt{KbsdJames/Omni-Judge}) and OpenAI's GPT-5 mini (\texttt{gpt-5-mini-2025-08-07}). 

\begin{figure}
    \centering
    \includegraphics[width=\textwidth]{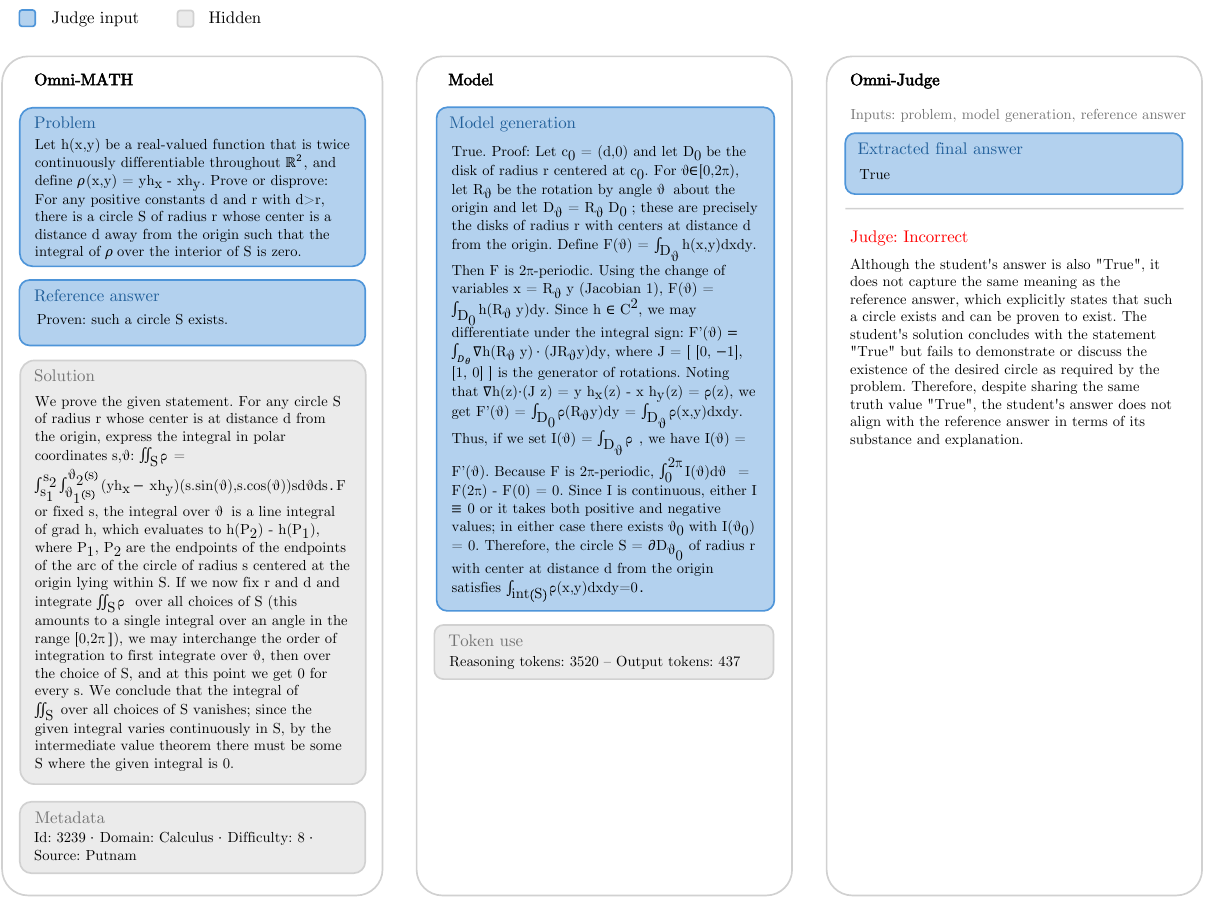}
    \caption{Example of the evaluation process for a question labelled as \emph{proof}.}
    \label{fig:ex3}
\end{figure}

\begin{figure}
    \centering
    \includegraphics[width=\textwidth]{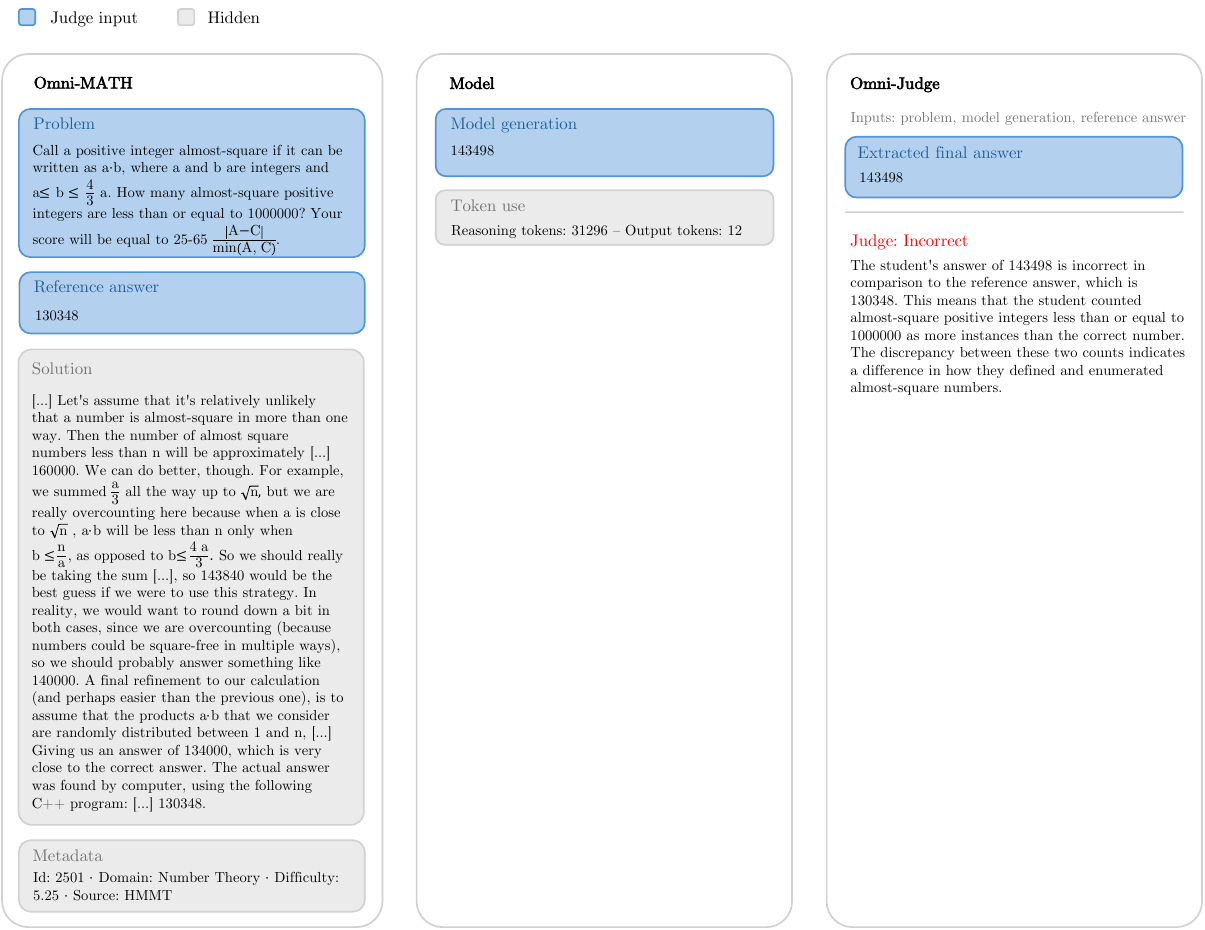}
    \caption{Example of the evaluation process for a question labelled as \emph{estimation}.}
    \label{fig:ex4}
\end{figure}

\begin{figure}
    \centering
    \includegraphics[width=\textwidth]{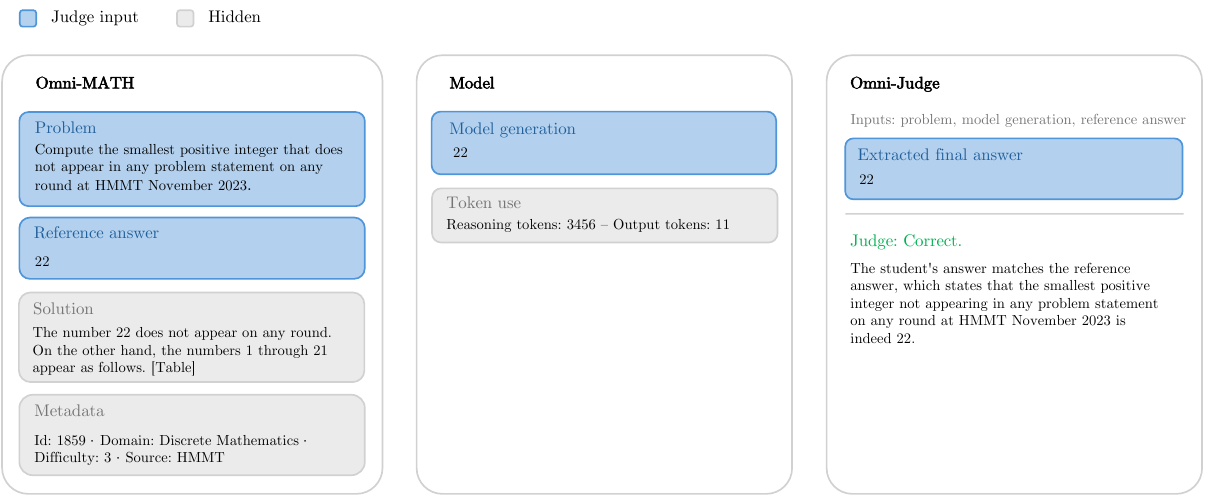}
    \caption{Example of the evaluation process for a question labelled as \emph{should delete}.}
    \label{fig:ex1}
\end{figure}

\paragraph{Omni-Judge} Omni-Judge is an efficient and low cost open-source math-evaluation model developed by the authors of Omni-MATH \cite{gao2024omni}. The model is trained to assess the correctness of an answer generated by an LLM, given the problem and a reference answer (see \Cref{fig:ex1,fig:ex2,fig:ex3,fig:ex4}). We make requests to the chat completions endpoint of the \texttt{kbsdjames.omni-judge} API  by running the model in LM Studio. We use the same few-shot prompt as in \cite{gao2024omni} and set the \texttt{max\_new\_tokens} parameter to $300$, and context length to $4{,}096$. 

\paragraph{GPT-5 mini} GPT-5 mini is a faster, more cost-efficient version of GPT-5, designed for well-defined tasks and precise prompts. In Humanities Last Exam \cite{phan2025humanity}, they use o3-mini for evaluating language models on frontier math, physics and other science problems. We adopt their evaluation prompt for GPT-5 mini, where we excluded the instruction to give a confidence score.
We ask the model to provide its answer in a json schema with fields extracted final answer, reasoning, and correct. 

\clearpage

\clearpage

\clearpage

\clearpage

\begin{figure}
    \centering
    \includegraphics[width=\linewidth]{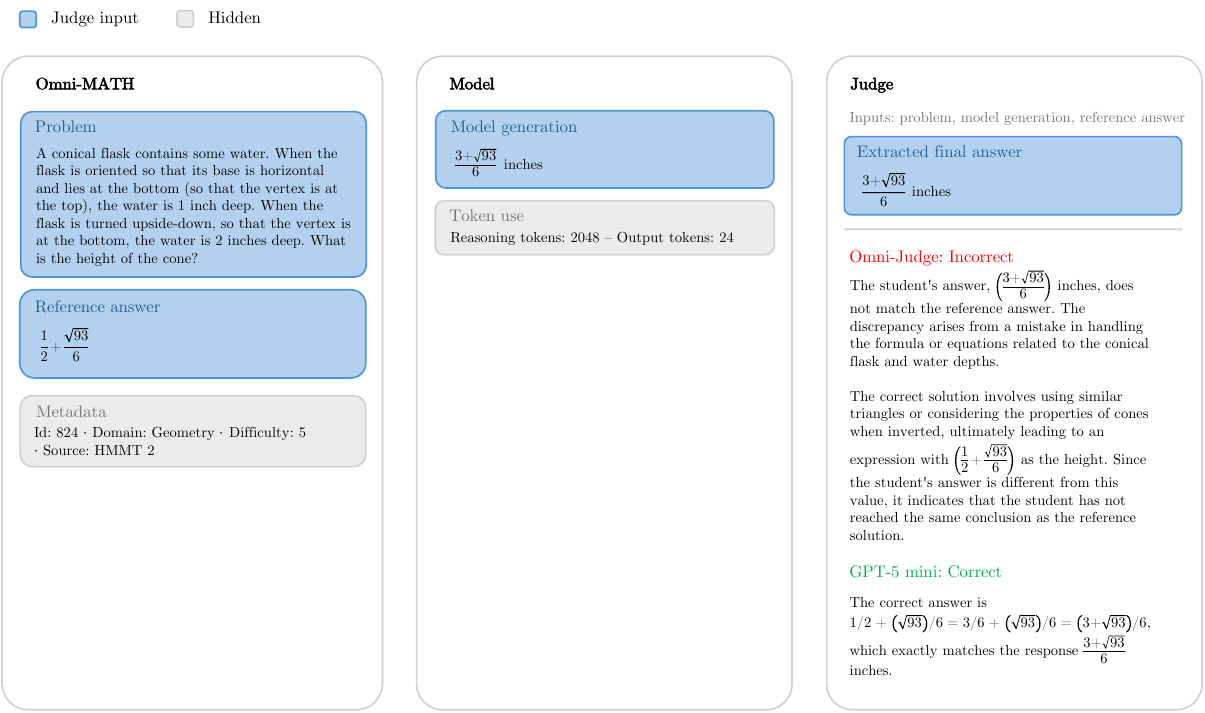}
    \caption{Example of a judge disagreement between Omni-Judge and GPT-5 mini. Omni-Judge fails to assess that the fraction in the reference answer is equal to the fraction in the model's final answer.}
    \label{fig:easy_eq}
\end{figure}

\clearpage

\begin{figure}
    \centering
    \includegraphics[width=\linewidth]{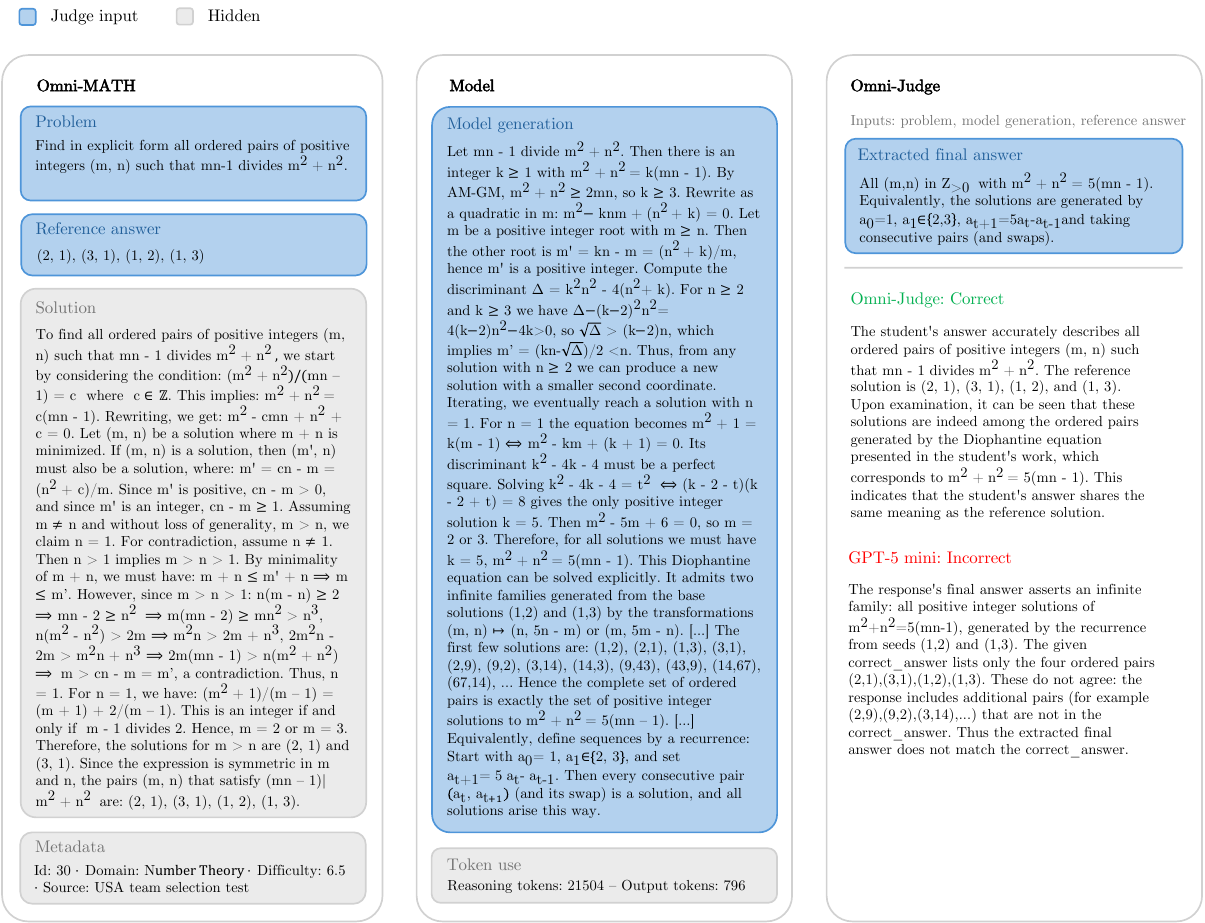}
    \caption{Example of a judge disagreement between Omni-Judge and GPT-5 mini. The reference answer is incomplete, causing GPT-5 mini to judge the model's answer as wrong, while it is actually the correct and complete answer.}
    \label{fig:incomplete-ref}
\end{figure}

\clearpage

\begin{table}[]
    \centering
    \begin{tabular}{L{4.5cm} r r r r r}
          & \textit{GPT-5 mini} & \textit{Omni-Judge} & \\ \addlinespace \addlinespace \addlinespace
        
        Claude Sonnet 4.5 & $79.29$ \small[78.03,80.49] & $74.93$ \small[73.65,76.28] & \\ \addlinespace
        
        DeepSeek v3.2 & $82.95$ \small[81.78,84.06] & $77.45$ \small[76.16,78.69]\\ \addlinespace
                
        Gemini 3 Pro (preview) & $89.93$ \small[88.98,90.81] &  $83.43$ \small[82.29,84.53] \\ \addlinespace
                GPT-5 & $84.53$ \small [83.40,85.59]   & $83.47$  \small[82.35,84.60] &  \\  \addlinespace
        Kimi K2 Thinking & $86.87$ \small[85.81, 87.86] &  $86.99$\small [86.00,88.03]\\ \addlinespace

        \addlinespace \addlinespace
    \end{tabular}
    \caption{Accuracy of Claude 4.5 Sonnet, DeepSeek v3.2, Gemini 3, GPT-5 and Kimi K2 Thinking on Omni-MATH-2-Filtered with Bayesian iid confidence intervals \cite{bowyer2025position}.}
    \label{tab:Table3}
\end{table}

\begin{table}[t]
    \centering
    \begin{tabular}{L{4.5cm} r r r r r r}
     & Total wrong & \multirow{2}{6.1em}{Failed to assess equivalence} & \multirow{2}{5.3em}{Didn't follow instructions} &  \multirow{2}{3em}{Dataset error} &  \multirow{2}{3em}{Wrong extraction} &  Unclear \\ 
     \\ \addlinespace \addlinespace \addlinespace
     Judge disagreements (n=100): \\  \addlinespace \addlinespace
         \emph{Omni-Judge} &  86 $\pm$ 1 & 75 & 10 & 5 & 2 & 0\\ \addlinespace
         \emph{GPT-5 mini} &  12 $\pm$ 1 & 1  & 1 & 9 & 0 & 1 \\ \addlinespace \addlinespace \addlinespace

    \end{tabular}
    \caption{Excluding the dataset errors, Omni-Judge is wrong in 96.4$\%$ of the subsampled judge disagreements with GPT-5 mini for Omni-MATH-2-Filtered. This is mainly because Omni-Judge is not able to assess equivalence between the model's final answer and the reference answer. } 
    \label{tab:clean disagreements}
\end{table}

\end{document}